\documentclass[10pt,twocolumn,letterpaper]{article}

\usepackage{cvpr}
\usepackage{times}
\usepackage{graphicx}
\usepackage{amsmath}
\usepackage{amssymb}

\usepackage{color,soul}
\usepackage{caption}
\usepackage{subcaption}
\usepackage{multirow}
\usepackage{xcolor,colortbl}
\definecolor{Gray}{gray}{0.95}
\DeclareMathOperator*{\median}{median}

\usepackage[pagebackref=true,breaklinks=true,letterpaper=true,colorlinks,bookmarks=false]{hyperref}

\cvprfinalcopy 

\ifcvprfinal\fi
\begin{document}

\title{Scene Invariant Crowd Segmentation and Counting Using Scale-Normalized Histogram of Moving Gradients (HoMG)}

\author{Parthipan Siva\\
Aimetis Corp.\\
\and
Mohammad Javad Shafiee\\
University of Waterloo\\
\and
Mike Jamieson \\
Aimetis Corp.\\
\and
Alexander Wong \\
University of Waterloo
}

\maketitle

\begin{abstract}
The problem of automated crowd segmentation and counting has garnered significant interest in the field of video surveillance. This paper proposes a novel scene invariant crowd segmentation and counting algorithm designed with high accuracy yet low computational complexity in mind, which is key for widespread industrial adoption.  A novel low-complexity, scale-normalized feature called Histogram of Moving Gradients (HoMG) is introduced for highly effective spatiotemporal representation of individuals and crowds within a video.  Real-time crowd segmentation is achieved via boosted cascade of weak classifiers based on sliding-window HoMG features, while linear SVM regression of crowd-region HoMG features is employed for real-time crowd counting.  Experimental results using multi-camera crowd datasets show that the proposed algorithm significantly outperform state-of-the-art crowd counting algorithms, as well as achieve very promising crowd segmentation results, thus demonstrating the efficacy of the proposed method for highly-accurate, real-time video-driven crowd analysis.
\end{abstract}

\section{Introduction}
Given the ever-increasing demands for video-driven monitoring of busy public spaces, there has been significant interest in determining the presence and distributions of crowds of people in an automated fashion.  Such automated monitoring and analysis of crowds has numerous applications ranging from analyzing crowd congestion patterns and customer attention behaviour, to detecting long queues, unsafe crowding or even mass panic.  In particular, the crowd segmentation and counting problem involves the localization of crowds of people within a scene, and estimating the number of individuals within such crowds.  Tackling this problem is particularly important for busy scenes, where detecting and tracking individuals in the video can be both computationally expensive and unreliable.

This ever-growing interest in automated crowd analysis has prompted researchers to propose a variety of methods for video-driven crowd segmentation and counting \cite{DBLP:journals/corr/KangW14b, Lin2011, Manfredi201439, Ryan201498, Zhang_2015_CVPR, ChanVasconcelos2012, MDE, Ghidoni2013, DBLP:journals/jsw/YangBWL13}, with a recent survey on state-of-the-art methods found in~\cite{Ryan20151}.  However, widespread industrial adoption of automated crowd segmentation and counting requires a simple, low-cost, highly-scalable deployment process, which is difficult to achieve using existing methods. Most existing methods require expensive, manually-annotated training data for each camera view, many scene-invariant methods are not designed for crowd counting (\eg \cite{Manfredi201439, DBLP:journals/corr/KangW14b}), and even methods that can perform per-camera training over a period of time (\cite{Zhang_2015_CVPR, Lin2011}) significantly complicate large-scale deployment given the need to adopt to different camera settings and views.  Ryan \etal \cite{Ryan2011, ryan2012, Ryan201498} have proposed scene-invariant approaches that do not require per-camera annotation or training, which is very important for large-scale deployment since the camera views of different cameras can be significantly varied.

In this paper, we propose a novel method for scene-invariant crowd segmentation and counting.  The proposed method centers around a novel, scale-normalized feature called Histogram of Moving Gradients (HoMG) that is designed to represent crowds with high accuracy yet low computational complexity. We show that our method significantly improves on the state-of-the-art for the seven videos tested in \cite{Ryan201498} and introduce an expanded set of annotated videos for testing.  While most crowd counting methods that rely on the camera's position relative to the ground plane to weight features in the regression to account for significant variations in the projected size of a person between the near and far range of most surveillance scenes, the proposed method accounts for scale adaptively, which is important as it is widely-recognized that features should be extracted in relation to the scale of the object to be detected (e.g., \cite{HOG_2005_Dalal}). 

\section{Methodology}

\begin{figure}[t]
\begin{center}
\includegraphics[width=0.9\linewidth]{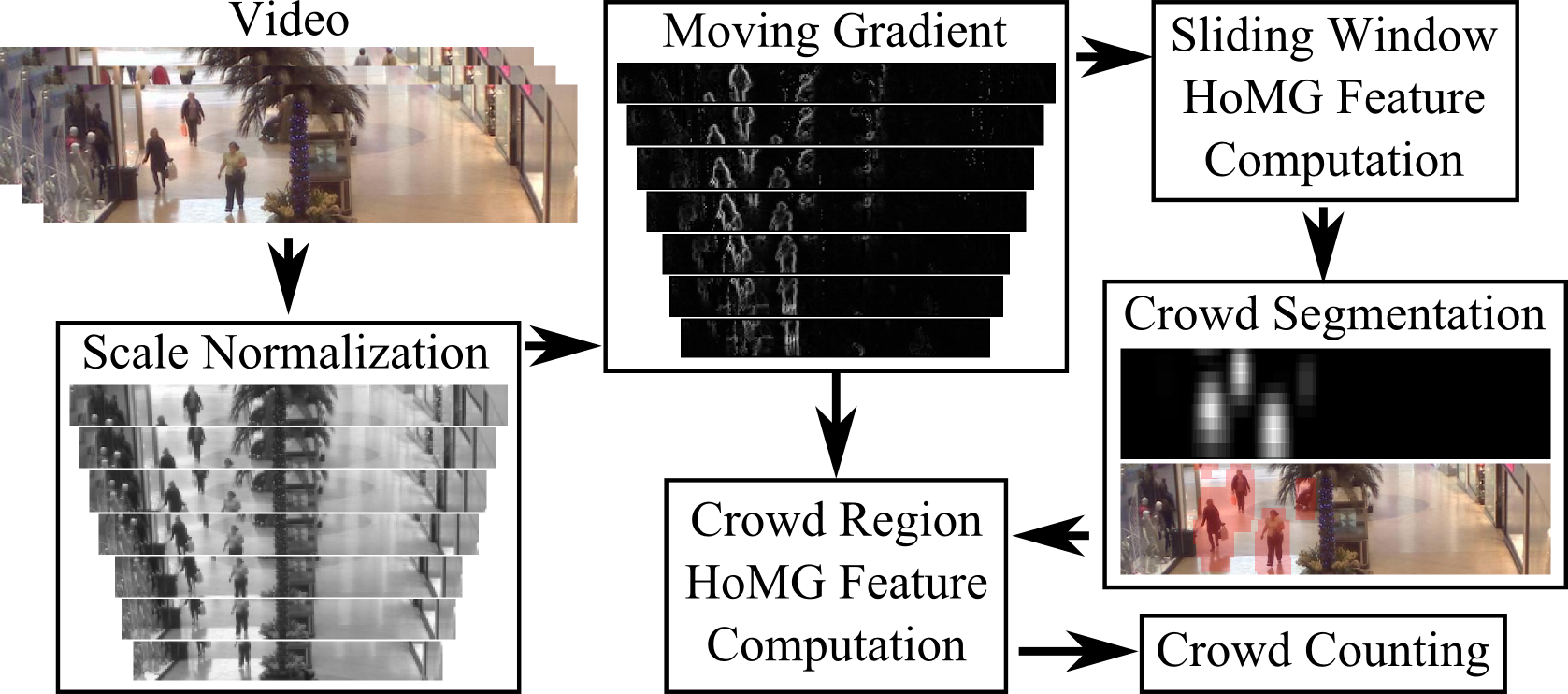}
\end{center}
   \caption{\small \em Overview of proposed method. First, scale-normalized moving gradients are computed. Second, sliding-window Histogram of Moving Gradient (HoMG) features are computed and used to segment crowd regions. Finally, crowd-region HoMG features are computed and used to obtain the crowd count.}
\label{fig:block}
\end{figure}

 An overview of the proposed crowd segmentation and counting method is shown in Fig.~\ref{fig:block}. First, scale-normalized moving gradients are obtained to facilitate the computation of HoMG features (Section~\ref{subsec:HoMG}). Based on the obtained moving gradients, sliding-window HoMG features are computed and classified to obtain crowd segments (Section~\ref{subsec:crowdSeg}). Finally, using both the crowd segmentation results and the scale normalized moving gradients, crowd-region HoMG features are computed and used by a linear regressor to obtain the crowd count (Section~\ref{subsec:crowdCounting}).

\subsection{Histogram of Moving Gradients (HoMG)}
\label{subsec:HoMG}
We introduce the concept of histogram of moving gradients (HoMG) a single, powerful, yet low-complexity, scale-normalized feature descriptor used for both crowd segmentation and counting.  We will show that sliding-window HoMG features can be used for crowd segmentation in an effective and efficient manner.  Furthermore, we show that the cumulative scale-normalized moving gradient magnitude in crowd regions is linearly related to the number of people in the frame, thus leading to crowd-region HoMG features that can be easily used to obtain the crowd count.
\subsubsection{Scale Normalization}
\label{subsubsec:scaleNorm}
A general approach to scale normalization is to compute the feature descriptors at a fixed scale. For example, in SIFT~\cite{SIFT}, descriptors are computed at a fixed window size based on the scale of the detected key points. Since our interests lie in crowd detection, we instead compute HoMG feature descriptors at a fixed scale relative to person size.  Therefore, in this work, the goal is to re-scale the expected person size windows at all locations in the image such that the width of the person is $w_p$ pixels wide.

\begin{figure}[t]
\begin{center}
\includegraphics[width=\linewidth]{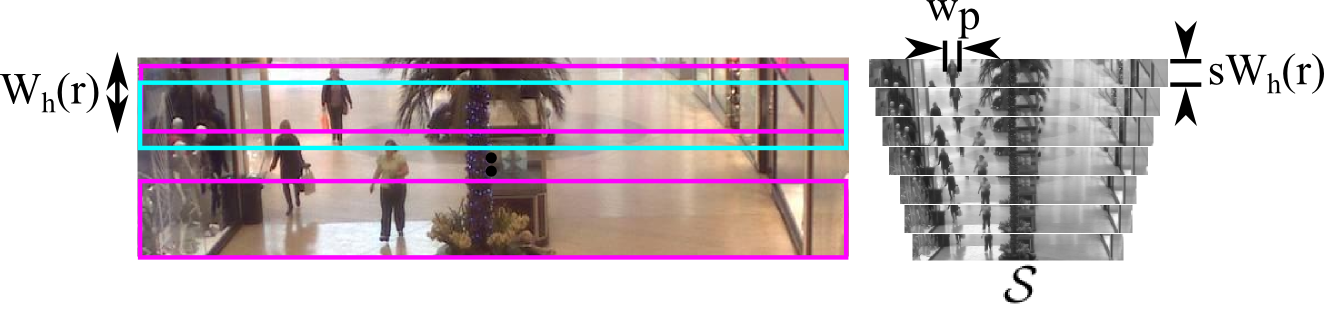}
\end{center}
   \caption{\small \em Scale normalization: each video frame is broken into overlapping strips of height $W_h(r)$ (the height of person at a given image row $r$) and re-scaled such that the person width in each strip goes to $w_p$ pixels.}
\label{fig:scaleNorm}
\end{figure}

The scale normalization process here is driven by the notion that while the general shape of people is the same, there can be significant clothing variations. As such, it is important to fix the scale such that it is small enough that clothing details are removed, yet large enough that the general shape of individuals are preserved.  Motivated by this, we leverage the camera calibration model, with the assumption that the person height and width per row is constant\footnote{while this assumption may not be very accurate due to lens distortion and camera rotations, such an assumption makes the sliding-window approach computationally tractable.}, and decompose the image into overlapping strips (Fig.~\ref{fig:scaleNorm}) of height $W_h(r)$, where $W_h(r)$ is the height of a person at row $r$.

Based on the calibration model, a person within each strip will have an average person width of $W_w(r)$ pixels. Therefore, each strip is re-scaled by $s$ such that the average person width becomes $w_p$ pixels (i.e., $s=\frac{w_p}{W_w(r)}$). The resulting scale-normalized frame $\mathcal{S}$ is illustrated in Fig.~\ref{fig:scaleNorm}.  Furthermore, based on the calibration model, any locations in the scene where the person size is less than $w_s$ pixels in width (in the original video resolution) will not be analyzed.

\subsubsection{Moving Gradients}
\label{subsubsec:movingGradient}
A key observation when incorporating temporal information for crowd analysis is that a person, even if they are waiting at a location, is never perfectly stationary over a period of time.  Such slight motions can cause problems for standard background subtraction techniques~\cite{MOG}, as parts of the person will become background while other parts are treated as foreground, thus leading to unreliable crowd segmentation.  To mitigate such issues, we identify moving gradients instead, allowing us to capture the outlines of individuals who are standing at a single location but exhibit slight motions.

For a given scale-normalized frame $\mathcal{S}$, we wish to identify its moving gradients, which essentially characterize the edges of moving foreground objects.  There are a number of approaches to computing moving gradients such as background modeling followed by edge detection, background modeling on the edge image, frame differencing, etc.  For the purposes of this paper, we chose to compute moving gradients based on frame differencing in the Sobel \cite{IP_2009_Gonzales} gradient domain as it is computationally efficient and allows us direct control of the temporal window to be considered.

Given an input video, we represent a scale-normalized grayscale strip at time $t$ as $S^{t} \in \mathcal{S}$, where $t=0$ refers the current strip and $t=-a$ refers to the strip $a$ seconds ago. Furthermore, we define the Sobel \cite{IP_2009_Gonzales} gradient of strip $S^{t}$ as $\vec{E}^t = (E_x^{t},E_y^{t})$, where $E_x$ is the horizontal gradient component and $E_y$ is the vertical gradient component.

\begin{figure}[t]
\begin{center}
\includegraphics[width=\linewidth]{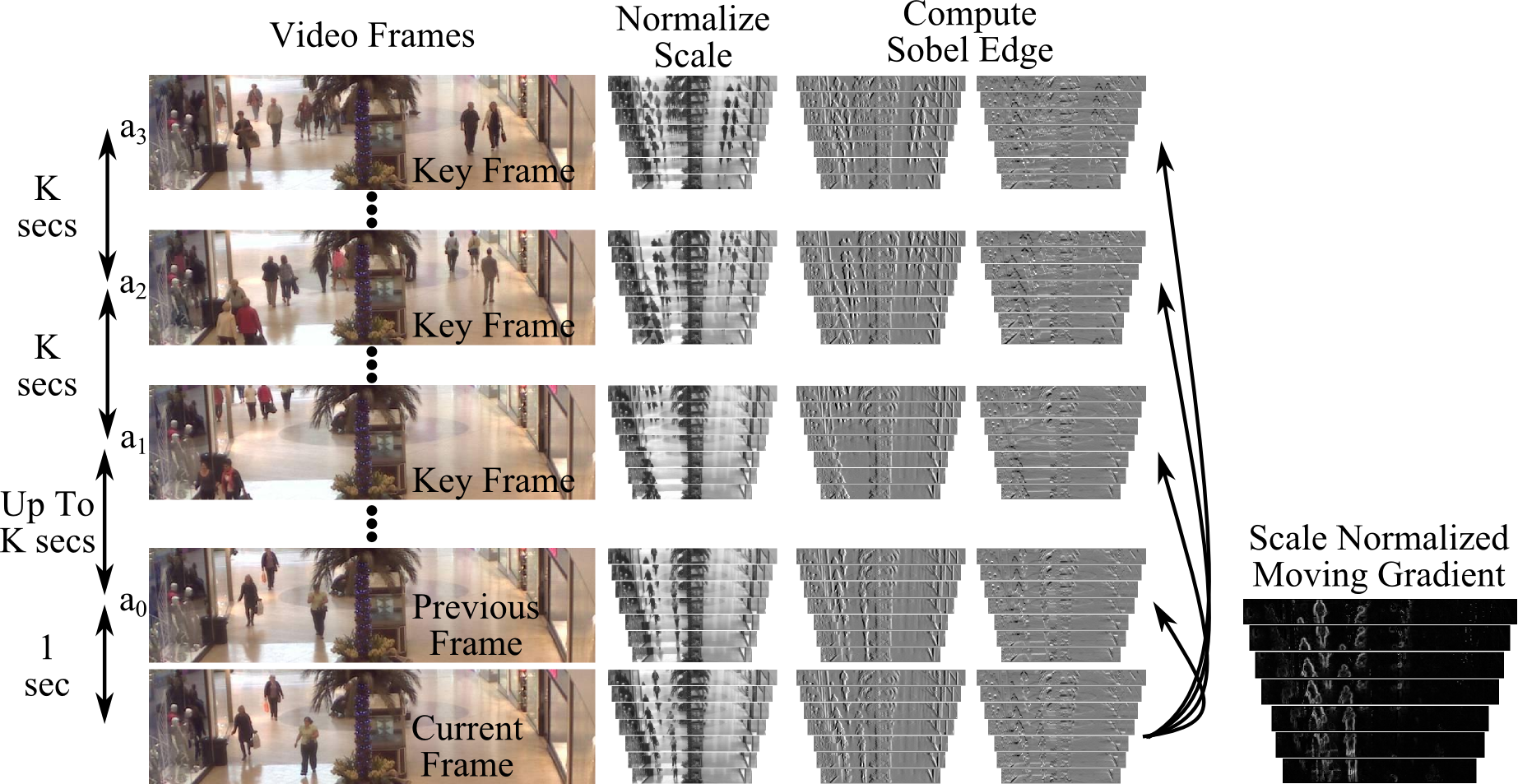}
\end{center}
   \caption{\small \em The moving gradient is obtained as the truncated median gradient between the current frame and a set consisting of the past frame and $l$ past keyframes.  The closest previous keyframe ($a_1$) can be up to $K$ seconds from the current frame.}
\label{fig:MovingGrad}
\end{figure}

The moving gradient, $E$, of a strip on the current frame $S^0$ is defined as the truncated median gradient between $S^0$ and a set $\mathcal{A}$, consisting of the past frame ($a_0$) and $l$ past keyframes ($a_1,\ldots,a_l$) (Fig.~\ref{fig:MovingGrad}):

{\footnotesize
\begin{eqnarray}
E &=& \min(\max(0,\bar{E}-T_1),T_2) \label{eq:movingEdge} \\
\bar{E} &=& \median_{a \in \mathcal{A}}(D^{a}) \label{eq:medianOfEdge}\\
\label{eq:edge_diff}
D^a &=& \left\{\def\arraystretch{1.2}%
  \begin{array}{@{}c@{\quad}l@{}}
    \|\vec{E}^0 - \vec{E}^a\| & \text{if $\|\vec{E}^0\|>\|\vec{E}^a\|$}\\
    0 & \text{else}\\
  \end{array}\right.
\end{eqnarray}
}

\noindent where $T_1$ and $T_2$ are thresholds on gradient magnitude to minimize the effects of very strong or very weak gradient magnitudes, and $\mathcal{A} = \{-a_0,-a_1,\ldots,-a_l\}$ is the temporal scale of the moving gradient. Eq.~\ref{eq:edge_diff} effectively detects edges on the current strip not present in the strip $a$ seconds ago. 

The moving gradient magnitude $E$ is given by Eq.~\ref{eq:movingEdge}, while the orientation $O$ is obtained as

{\footnotesize
\begin{equation}
\label{eq:movingEdgeOri}
O = \Bigl\lfloor N\frac{atan2\left(E^0_y,E^0_x\right)+\pi}{2\pi} \Bigr\rfloor
\end{equation}
}

\noindent where $\vec{E}^0$ denotes the Sobel gradient of the current frame, and $N$ represents the number of discretized orientation bins.

To allow for immediate reporting of crowd information when the algorithm is first started, we allow $\mathcal{A}$ to be small (i.e. $\mathcal{A} = \{-a_0\}$) and then grow to the full size $\mathcal{A} = \{-a_0,-a_1,\ldots,-a_l\}$.  As such, at least $a_0$ seconds of video is needed before reporting crowd statistics.

We only leverage a set consisting of the past frame ($a_0$) and past $l$ keyframes ($a_0,\ldots,a_l$) in Eq.~\ref{eq:medianOfEdge} rather than a large set of past frames (Fig.~\ref{fig:MovingGrad}).  Hence, only $l+1$ frames are need to be kept in memory and thus significantly reducing memory overhead.

\subsubsection{Histogram Construction}
\label{subsubsec:HoMG}
Based on the moving gradient magnitude $E$ and the moving gradient orientation $O$, the histogram of moving gradients (HoMG) within a window of interest $W$ is defined as

{\small
\begin{equation}
h(\theta) = \sum_{p \in W | O(p) = \theta} E(p)
\end{equation}
}

\subsection{Crowd Segmentation}
\label{subsec:crowdSeg}
\begin{figure}[t]
\begin{center}
\includegraphics[width=\linewidth]{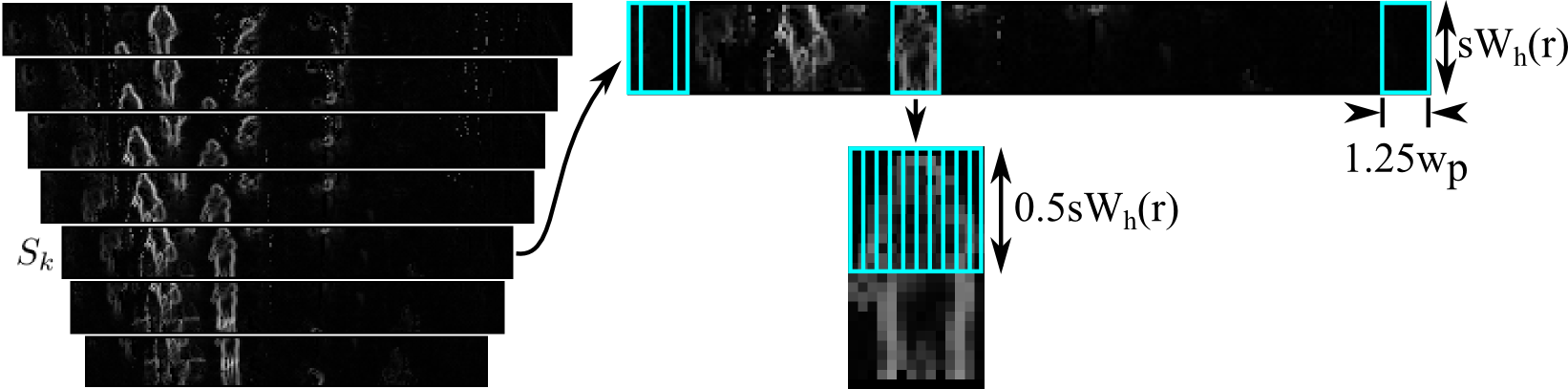}
\end{center}
   \caption{\small \em Illustration of a sliding window within a strip and the $1\times1.25w_p$ celled HoMG block used within a sliding window.}
\label{fig:slidingwindow}
\end{figure}

For crowd segmentation, we wish to detect the region occupied by crowds on each scale-normalized strip $S_k$ in $\mathcal{S}$.  This is accomplished using a sliding-window approach (Fig.~\ref{fig:slidingwindow}), with the window spanning the height of the strip and has a width of $1.25w_p$ pixels, where $w_p$ is the width of the person as defined in Section~\ref{subsubsec:scaleNorm}.

\subsubsection{Sliding-window HoMG}
Given a sliding window of interest, a single $1 \times 1.25w_p$ HoMG block is used as defined in Fig.~\ref{fig:slidingwindow}. The HoMG representation of the block is denoted by $h_{\theta}^g$, where $\theta$ is the discretized moving gradient orientations and $g$ is the cells in the HoMG block. Furthermore, we compute the total moving gradient (MG) magnitude within each of the cell as $s^g = \sum_{\theta}h_{\theta}^g$. Finally, the total moving gradient (MG) magnitudes within the entire window of interest $t$ is also computed. The HoMG block, MG magnitudes within the cells, and the MG magnitude within the entire windows are concatenated to form the crowd detection feature $f$.

{
\begin{equation}
\label{eq:detectionFeature}
f = \{h_{\theta}^g, s^g, t\}
\end{equation}
}

\noindent where $g$  is the cells in the HoMG block and $\theta$ is the discretized moving gradient orientations. In this work, we use $8$ moving gradient orientations and $1 \times 1.25w_p = 1 \times 10$ cells in the HoMG block; as a result, the crowd segmentation feature vector $f$ is of dimensions $1 \times 91$.

While all strips are scale-normalized to obtain constant person width, the person height (i.e., height of the strip) may still vary within an image due to perspective effects. As a result, we account for height by normalizing our feature vector $f$ \eqref{eq:detectionFeature} by the height of the strip.
\subsubsection{Classification}
\label{subsubsec:weakClassifier}
Given the computed feature vector, the next step is to classify whether the window of interest is crowd or non-crowd.  While there are numerous classification algorithms that can be used, our primary goal is to achieve a real-time algorithm and as such a boosted cascade of weak classifiers \cite{Cascade_2001_Viola} is used in this work for its low computational complexity at run time. Specifically, the AdaBoost \cite{Adaboost_1997_Freund} algorithm is used to obtain 100 boosted weak classifiers, where each weak classifier is a decision tree \cite{DT_2000_Dietterich} with three decision nodes.

For training, the positive samples centred around each annotated head location were used, with the negative samples obtained from unannotated areas. Since the negative space is much larger than the positive space, we employ the negative mining technique of \cite{negMine}.
\subsubsection{Segmentation}
\begin{figure}[t]
\begin{center}
\includegraphics[width=\linewidth]{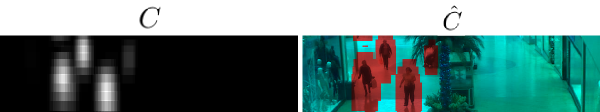}
\end{center}
   \caption{\small \em The accumulated crowd classification score image $C$ is thresholded to identify crowd regions $\hat{C}$ (shown in red).}
\label{fig:crowdImg}
\end{figure}

The boosted weak classifier is used to classify each sliding window on each strip as crowd or non-crowd with a confidence score. Each sliding window $W_s$ in the strip $S_k$ has a corresponding window $W$ in the original-resolution video frame. The scores are accumulated at the original video frame resolution using the correspondence window $W$. The accumulated score image $C$ is thresholded to obtain the crowd segmentation image $\hat{C}$ (Fig.~\ref{fig:crowdImg}).

\subsection{Crowd Counting}
\label{subsec:crowdCounting}
Here, we employ a linear regression approach to crowd counting using HoMG. We assume that the cumulative scale-normalized moving gradient magnitudes in the crowd segments are linearly related to the number of people in the video frame. While this is not strictly true due to occlusions and camera angles, we will show that this approach works very well in comparison to state-of-the-art.
\subsubsection{Crowd-region HoMG}
Based on the aforementioned linear relationship assumption, we introduce a cumulative HoMG for the crowd counting feature $R$, where only one histogram bin is used (i.e., moving gradient orientation is ignored):

{\footnotesize
\begin{equation}
R = \sum_{S_k \in \mathcal{S}} \frac{1}{h_k} \sum_{p \in S_k} \delta(p)E(p)
\end{equation}
}

\noindent where $S_k \in \mathcal{S}$ are all the scale-normalized strips in the frame, $h_k$ is the height of the $k^{\text{th}}$ strip after scale normalization, $p \in S_k$ are all the pixels in stripe $S_k$, $E(p)$ is the moving edge magnitude at pixel $p$, and $\delta(p) = \{0,1\}$ is the crowd detection result at pixel $p$.  This results in a single one-dimensional feature which is proportional to the number of people in the frame.
\subsubsection{Regression}
\label{subsubsec:regress}
We use linear SVR \cite{SVR_2004_Cherkassky}  as our regression method. Linear SVR was chosen for two reasons: i) its robust nature when fitting a line to the data, and ii) low computation complexity at run time. Furthermore, our assumption is that there is a linear relationship between moving gradient magnitude and the number of people in the video frame; as such it goes to reason that when the moving gradient magnitude approaches zero, the number of people in the frame approaches zero. As a result, we assume the bias term in our linear SVR model is zero. 

\section{Evaluation Setup}

To quantitatively evaluate the proposed crowd segmentation and counting method, which we will refer to as HoMG, we introduce a new multi-camera dataset, which has almost double the number of camera views as previously-used datasets, which is described below. Furthermore, competing methods used for evaluation as well as evaluation metrics used are also described below.

\subsection{Dataset}

Several crowd detection datasets are available in the literature \cite{Loy2013,Ryan201498}. However, they employ various different mechanisms for annotation, such as using the heads of people \cite{Loy2013} and the centers of people \cite{Ryan201498}, making them difficult to evaluate as a common benchmark.  Furthermore, there are also different datasets used for evaluating tracking \cite{benfold2009attention} and video surveillance \cite{iLIDs_2006_HOSDB} which could be used but are not designed for crowd analysis evaluation.  Motivated to create a unified benchmark dataset for evaluating crowd analysis, we introduce a new dataset for scene-invariant crowd counting which brings together existing crowd detection datasets \cite{Loy2013,Ryan201498}, tracking datasets \cite{benfold2009attention} as well as a number of new videos, all with a consistent annotation format.  Furthermore, for each camera view, we provide coarse camera calibration information as discussed in Section~\ref{subsec:algSetup}.

\begin{figure}
\begin{center}
\includegraphics[width = \linewidth]{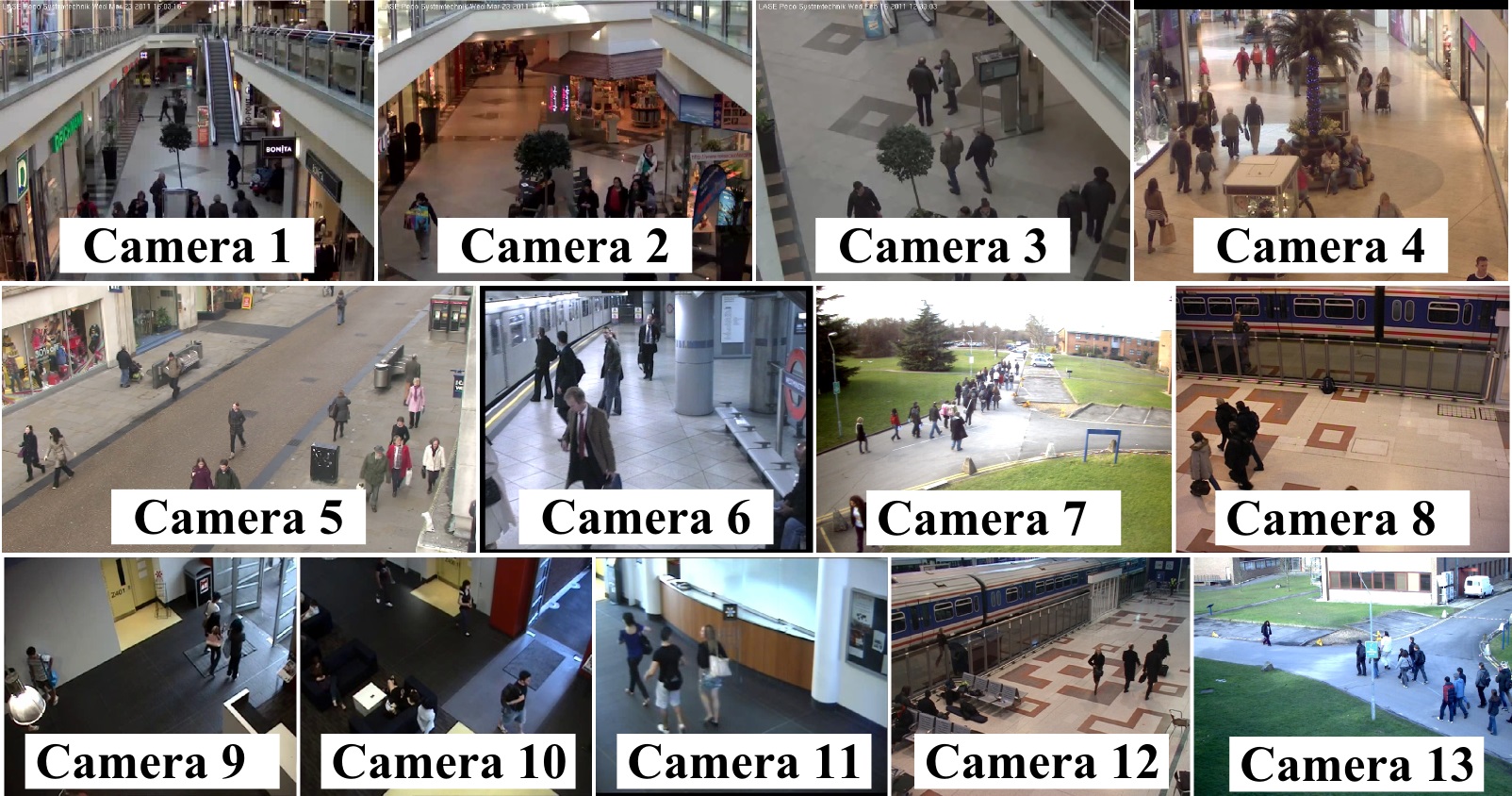}
\caption{\small \em Camera view examples for the proposed new dataset: comprising of standard datasets (camera 4-13) and 3 new camera views (camera 1-3).}
\label{fig:CameraExamples}
\end{center}
\end{figure}

The proposed dataset has 13 different camera views (Fig.~\ref{fig:CameraExamples}).
Cameras 1, 2 and 3 are the new video datasets being introduced by this paper for the first time. Camera 4 is the Mall dataset introduced in~\cite{Loy2013}. Camera 5 is the City Center dataset~\cite{benfold2009attention} which have been used for visual tracking problems. Camera 6 was selected from i-LIDS video datasets~\cite{iLIDs_2006_HOSDB}. Finally, cameras 7-13 are PETS and QUT video datasets used by Ryan {\it et al.}~\cite{Ryan201498}.  A subset of the dataset which only includes camera views 7-13 are also used to analyze the proposed HoMG method to facilitate for direct comparisons to published results \cite{Ryan201498}.

The training and testing procedure are done based on a leave-one-out framework (same as~\cite{Ryan201498}) where one camera view is assigned as the testing view, while the tested methods are trained with the remaining camera views. Since each camera view has a different number of annotated frames, data balancing is required during training to ensure equal weight are assigned to all camera views. As a result, only the first 50 annotated frames from each camera view are used for training.

\subsection{Competing Methods}
\label{subsec:comparisons}
The proposed HoMG method is compared quantitatively with two state-of-the-art frameworks:

\noindent{\bf Single Camera Crowd Counting (S3C)} Several types of features including segments, edges, GLCM, and LBP are extracted from video sequence, and a kernel ridge regression (KRR)~\cite{Loy2013} is utilized to estimate the crowd count of the scene. This method originally was used in the situation when training and test videos are from the same camera view. However, this method is compared here in a scene-invariant scenario.  \\
\noindent{\bf Scene Invariant Multi Camera Crowd Counting (SIM3C)} This method~\cite{Ryan201498} is the state-of-the-art approach in scene invariant crowd counting that doesn't require any adaptation to the camera view being tested. SIM3C obtains several types of features including size, shape, edge, and keypoints extracted using SURF~\cite{SURF}, and feeds these features into a non-linear Gaussian process regression framework to estimate the crowd count.

\subsection{Evaluation Metric}
Crowd counting performance is evaluated using three different quantitative metrics: Mean Absolute Error (MAE)~\cite{Loy2013}, Mean Square Error (MSE)~\cite{Loy2013} and Mean Deviation Error (MDE)~\cite{MDE}.

\noindent\begin{minipage}{.5\linewidth}
{\footnotesize
\begin{equation}
MAE = \frac{1}{N} \sum_{i =1 }^N |y_n - \hat{y}_n|
\end{equation}
}
\end{minipage}%
\begin{minipage}{.5\linewidth}
{\footnotesize
\begin{equation}
MSE = \frac{1}{N} \sum_{i =1 }^N (y_n - \hat{y}_n)^2
\end{equation}
}
\end{minipage}
\vspace{-0.7cm}
\begin{center}
\begin{minipage}{.5\linewidth}
{\footnotesize
\begin{equation}
MDE = \frac{1}{N} \sum_{i =1 }^N \frac{|y_n - \hat{y}_n|}{y_n}
\end{equation}
}
\end{minipage}
\end{center}

\noindent where $N$ represents the number of testing frames, and $y_n$ and $\hat{y}_n$ denote the actual count and the estimated count of frame $n$, respectively.

Crowd segmentation performance is evaluated using precision-recall curves (as a function of threshold on $C$) and average precision (AP) as defined by the PASCAL Segmentation challenge~\cite{pascal}. The ground truth segmentation is obtained based on the ground-truth head annotation of the dataset and the provided coarse camera calibration, where a binary mask is created by placing a rectangle of an average person size at each head locations.

\subsection{Algorithm Setup}
\label{subsec:algSetup}

As with other crowd counting methods \cite{Ryan201498,Loy2013}, a camera calibration model is needed for our HoMG method. In~\cite{Loy2013}, a very simple perspective model is used which does not account for the non-linear behaviour of perspective models. In \cite{Ryan201498}, a full calibration model is used; however, this is often unavailable in industry surveillance applications. Therefore, we instead employ a model currently employed in industry that acts as a compromise between \cite{Loy2013} and \cite{Ryan201498}.  Here, we use the interactive camera calibration model used commercially in the Aimetis Symphony~\cite{symphony} surveillance software package, where an approximate calibration model is obtained by having the users select calibration parameters that matches the model's estimated person sizes to the person sizes in the video.

Using this calibration model, we set the scaled person width ($w_p$), from Section~\ref{subsubsec:scaleNorm} to $w_p=8$ pixels. We select $8$ pixels as a value small enough to detect crowds far from camera but at the same time is large enough to learn features to describe individuals and crowds.

Similar to \cite{Ryan201498,Loy2013}, which requires parameters for background subtraction algorithms used to obtain moving blobs, the HoMG algorithm requires the temporal scale $\mathcal{A}=\{1,K,2K,\ldots,lK\}$ (Section~\ref{subsubsec:movingGradient}) to be defined for computing HoMG features.  A temporal scale of $120$ seconds was used to determine moving gradients, where keyframes are set $30$ seconds apart (i.e., $l=4$ and $K=30$). The scale of $120$ seconds was chosen as it is long enough to ensure people who exhibit slight motions be identified, while still being computationally tractable.

For crowd segmentation (Section~\ref{subsubsec:weakClassifier}), we use the AdaBoost classifier implemented in Matlab \cite{matlab}, with a $3$-node decision tree as the weak classifier.  All other learning parameters were left to the default values. For crowd counting (Section~\ref{subsubsec:regress}), we use the LIBLINEAR \cite{liblinear} implementation of linear SVR with default learning parameters.

\section{Crowd Segmentation Results}

\begin{figure}[t]
\begin{center}
\begin{minipage}{0.75\linewidth}
\includegraphics[width=\linewidth]{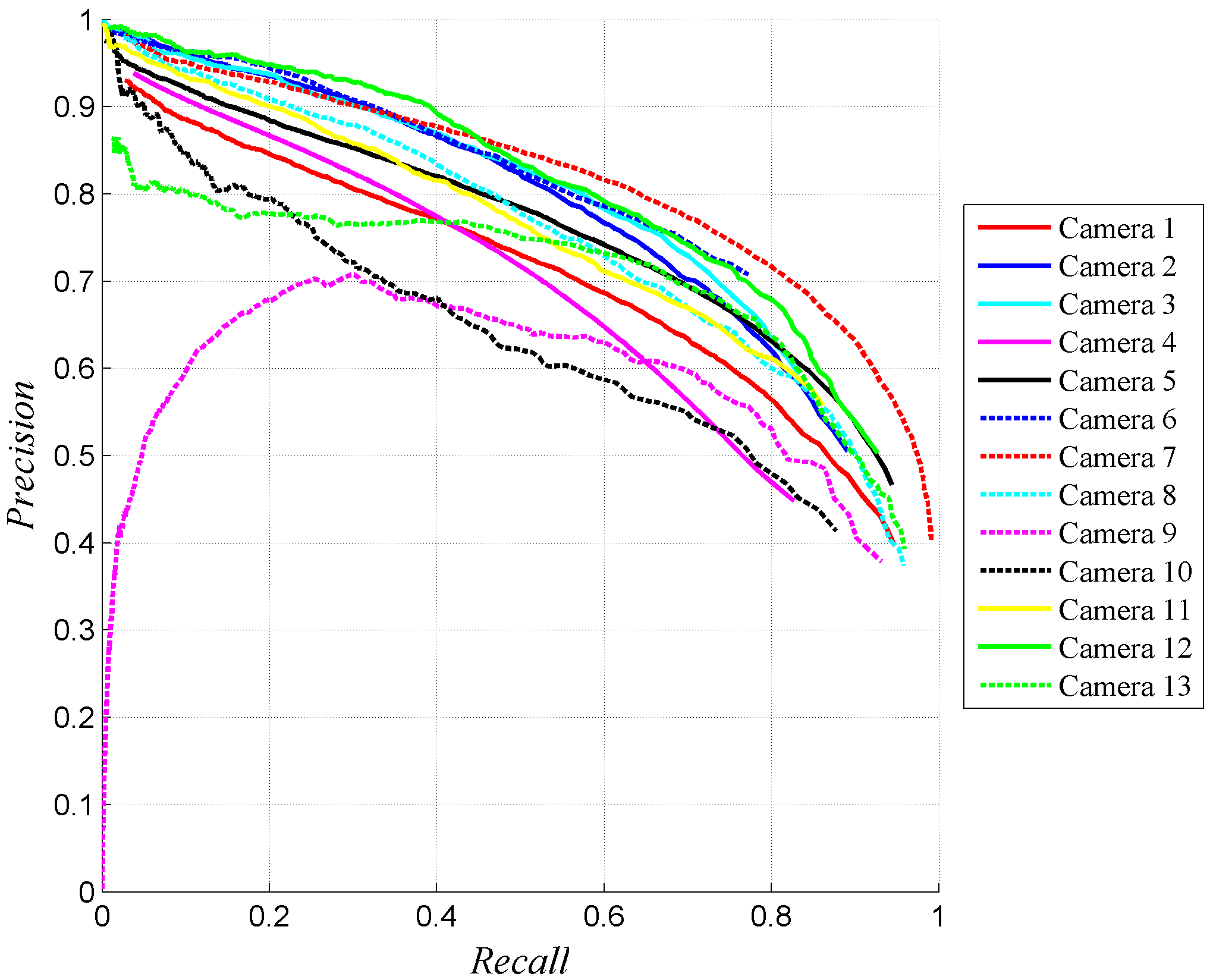}
\end{minipage}
\begin{minipage}{0.2\linewidth}
\setlength{\tabcolsep}{0.5mm}
\scriptsize{
\begin{tabular}{|c|c|}
\hline
\textbf{Camera} & \textbf{AP} \\
\hline
1 & 0.682 \\
2 & 0.727 \\
3 & 0.701 \\
4 & 0.612 \\
5 & 0.729 \\
6 & 0.668 \\
7 & 0.812 \\
8 & 0.734 \\
9 & 0.587 \\
10 & 0.583 \\
11 & 0.680 \\
12 & 0.768 \\
13 & 0.688 \\
\hline
\hline
\textbf{Average} & \textbf{0.690} \\
\hline
\end{tabular}
}
\end{minipage}
\end{center}
   \caption{\small \em The precision-recall curves (PRC) and average precision (AP) for leave-one-out crowd segmentation test using  HoMG.}
\label{fig:prc}
\end{figure}

The precision-recall curves (PRC) and average precision (AP) numbers for the leave-one-out testing on all 13 camera views are presented in Fig.~\ref{fig:prc}. Overall on the 13 camera views, HoMG achieved an average AP of $0.690$.

\begin{figure}[t]
\begin{center}
\includegraphics[width=0.3\linewidth]{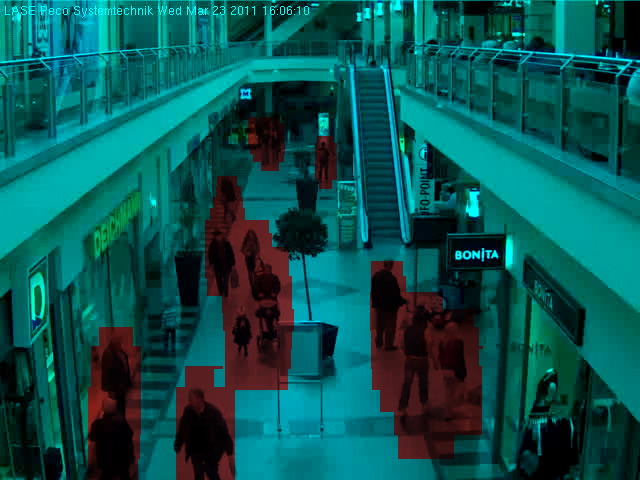}
\includegraphics[width=0.3\linewidth]{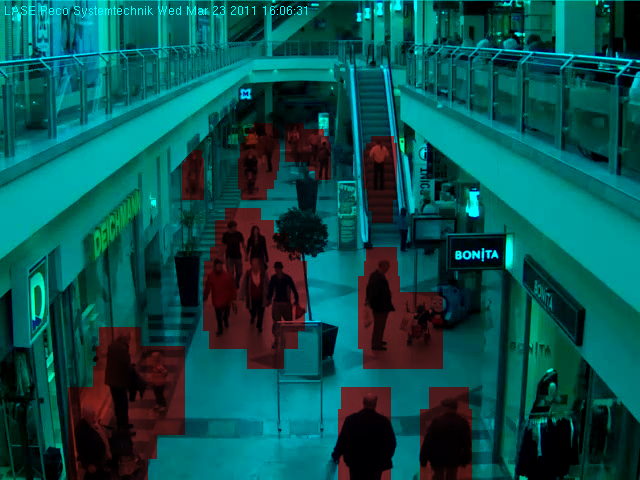}
\includegraphics[width=0.3\linewidth]{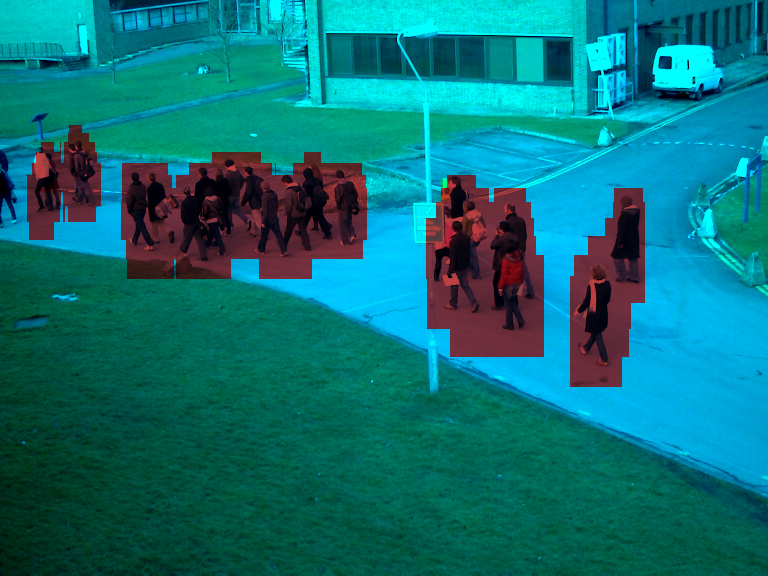}
\includegraphics[width=0.3\linewidth]{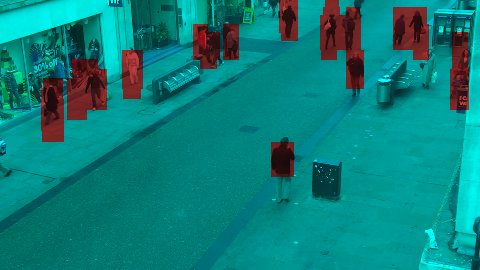}
\includegraphics[width=0.213\linewidth]{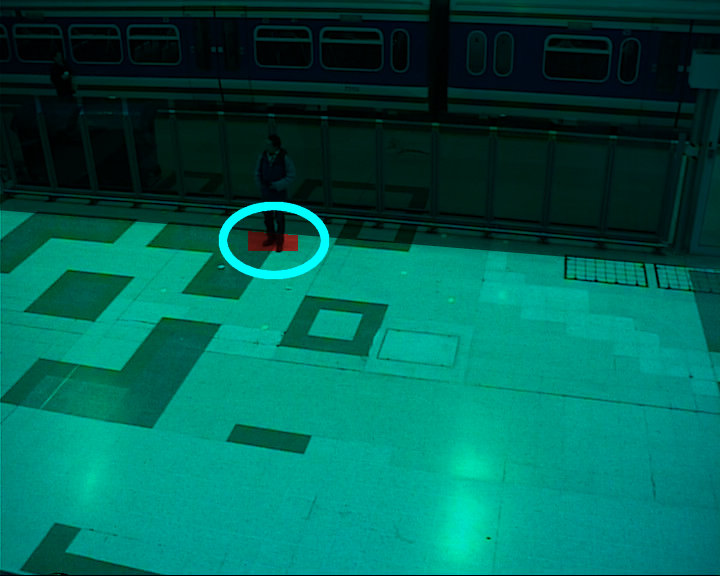}
\end{center}
   \caption{\small \em Example crowd segmentation results using HoMG. The overlay threshold is set at a value that maximizes F1-Score on all camera views. Bottom-right shows example of partial person detection due to the application of region-of-interest masks.}
\label{fig:exampleSeg}
\end{figure}

From Fig.~\ref{fig:prc}, it can be observed that the PRC for Camera 9 looks odd in comparison to that for the rest of the camera views. The reason for this anomaly is that in the evaluation process, crowd segments were computed in the full frame and then a mask indicating the region of interest is applied (same masks used in \cite{Ryan201498}). In this particular camera view (Fig.~\ref{fig:exampleSeg}, lower right) an individual is partially inside the mask but is not part of the ground-truth annotation. As a result, HoMG picks up the part of the person in the mask of interest which is considered a false positive, leading to a lowered precision value.  However, the AP results are not affected by this as the PRC is assumed to be monotonically decreasing when computing AP~\cite{pascal}.

Example crowd segmentation results are shown in Fig.~\ref{fig:exampleSeg} illustrating HoMG method's ability to detect both individuals and groups. Furthermore, background motion from lighting and background objects like escalators are ignored unless a person is found in the area, which is important for reliable crowd analysis.

\section{Crowd Counting Results}

Performance analysis of HoMG for crowd counting is carried out on the new 13 camera view dataset (Section~\ref{subsec:fullSetResults}).  Furthermore, to allow for direction comparison with state-of-the-art~\cite{Ryan201498}, HoMG is also tested on a 7 Camera subset of the dataset (Section~\ref{subsec:datasubset}), which contains only the camera views used by~\cite{Ryan201498}.

\subsection{Full Dataset}
\label{subsec:fullSetResults}
\begin{figure}[t]
\begin{center}
\includegraphics[width=\linewidth]{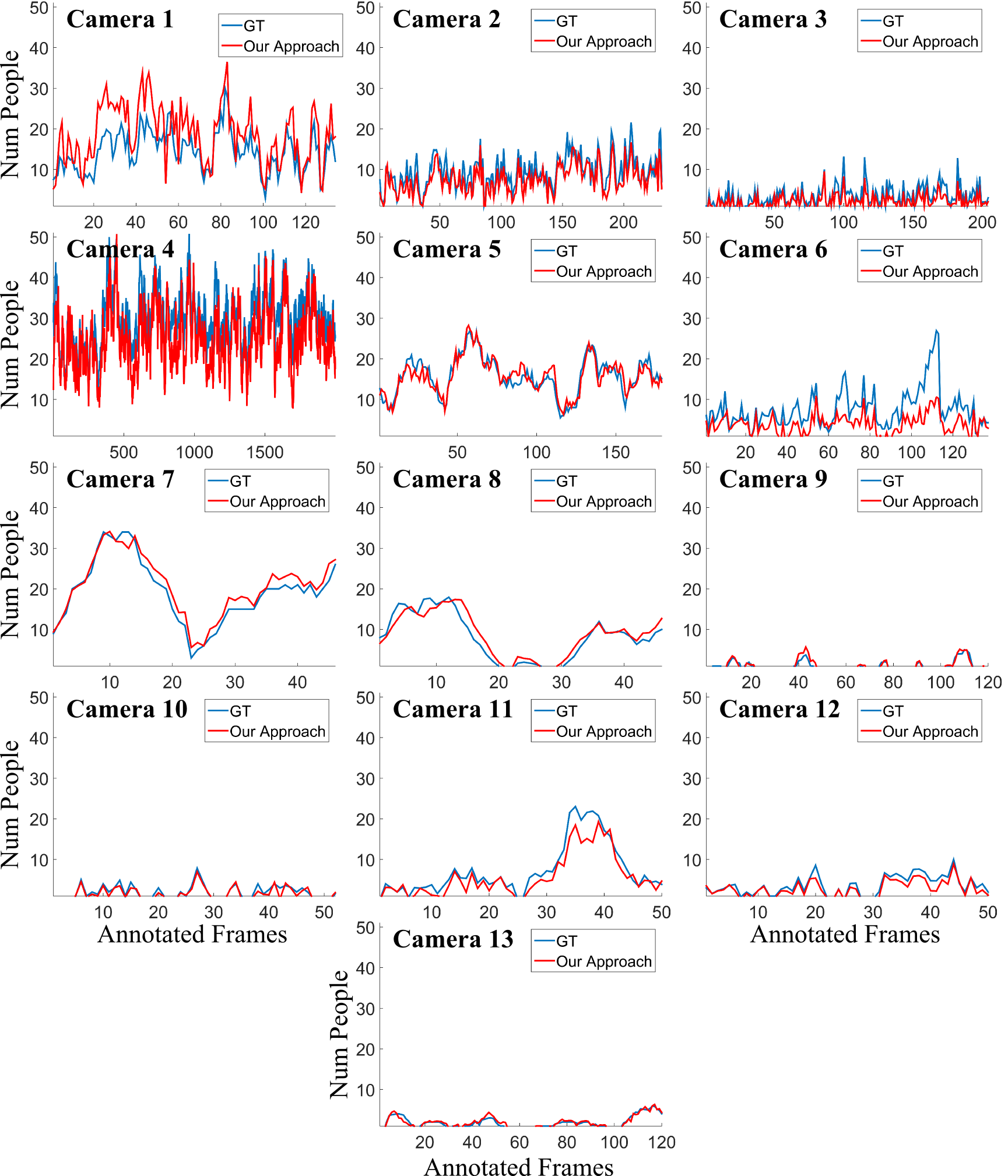}
\end{center}
   \caption{\small \em The ground truth (GT) and predicted counts for all annotated frames, from the leave-one-out testing.}
\label{fig:countPlots}
\end{figure}

\begin{table}
\begin{center}
\setlength{\tabcolsep}{0.5mm}
\scriptsize{
\begin{tabular}{|c|c|c|c|c|c|c|c|c|c|c|c|c|c|c|c|}
\hline
\multicolumn{2}{|c|}{\tiny{\textbf{Camera}}} & 1 & 2 & 3 & 4 & 5 & 6 & 7 & 8 & 9 & 10 & 11 & 12 & 13 & Avg. \\
\hline
 \rowcolor{Gray}
\multirow{2}{*}{\raisebox{-4 pt}[0 pt][0 pt]{\rotatebox{90}{\tiny \textbf{MAE}}}} & \tiny{HoMG} & \textbf{4.92} & \textbf{2.09} & \textbf{1.33} & \textbf{5.34} & \textbf{1.53} & \textbf{3.86} & \textbf{1.82} & \textbf{1.41} &\textbf{ 0.31} & \textbf{0.49} & \textbf{1.75} & \textbf{0.85} & \textbf{0.33} & \textbf{2.00} \\
&\tiny{S3C} & 15.0 & 17.9 & 7.9 & 23.7 & 15.5 & 4.0 & 9.1 & 7.8 & 1.5 & 2.5 & 2.8 & 1.7 & 1.6 & 7.9 \\\hline \hline
 \rowcolor{Gray}
\multirow{2}{*}{\raisebox{-4 pt}[0 pt][0 pt]{\rotatebox{90}{\tiny \textbf{MSE}}}} & \tiny{HoMG} & \textbf{35.1} & \textbf{6.9} & \textbf{3.0} & \textbf{38.8} &\textbf{ 3.7} & \textbf{24.4} & \textbf{1.7} &\textbf{ 3.5} & \textbf{0.3} &\textbf{ 0.5} & \textbf{6.2 }& \textbf{1.3} & \textbf{0.2} & \textbf{9.9} \\
&\tiny{S3C} & 62 & 339 & 83 & 608 & 234 & 25 & 99 & 95 & 4 & 10 & 15 & 5 & 4 & 122 \\\hline\hline
 \rowcolor{Gray}
\multirow{2}{*}{\raisebox{-4 pt}[0 pt][0 pt]{\rotatebox{90}{\tiny \textbf{MDE}}}} & \tiny{HoMG} & \textbf{0.35} & \textbf{0.23} & \textbf{0.37} & \textbf{0.18} & \textbf{0.11} & \textbf{0.49} &\textbf{ 0.12} & \textbf{0.25} &\textbf{ 0.35} &\textbf{ 0.28} & \textbf{0.26} & \textbf{0.24} & \textbf{0.20} & 0\textbf{.27} \\
 & \tiny{S3C} & 1.0 & 2.1 & 3.6 & 0.8& 0.9 & 0.9 & 0.5 & 3.3 & 47.8 & 203 & 0.5 & 0.7 & 1.7 & 20.5\\
 \hline
\end{tabular}
}
\caption{\small \em Crowd counting results on the full dataset}\label{tab:crowdCount13Cam}
\end{center}
\end{table}

The crowd counting results are shown in Table~\ref{tab:crowdCount13Cam} and Fig.~\ref{fig:countPlots}. The proposed HoMG method achieved relatively low average error across all camera views with respect to both MSE and MAE. In comparison to \textbf{S3C}, which is a baseline comparison, HoMG achieves almost four times lower MAE rate and over twelve times lower ASE rate.  However, as previously stated in Section~\ref{subsec:comparisons}, \textbf{S3C} was not originally designed as a scene invariant approach; nevertheless, we treat it as such because it does have a very coarse perspective normalization.

\begin{figure}[t]
\begin{center}
    \begin{subfigure}[b]{0.46\linewidth}
        \includegraphics[width=\textwidth]{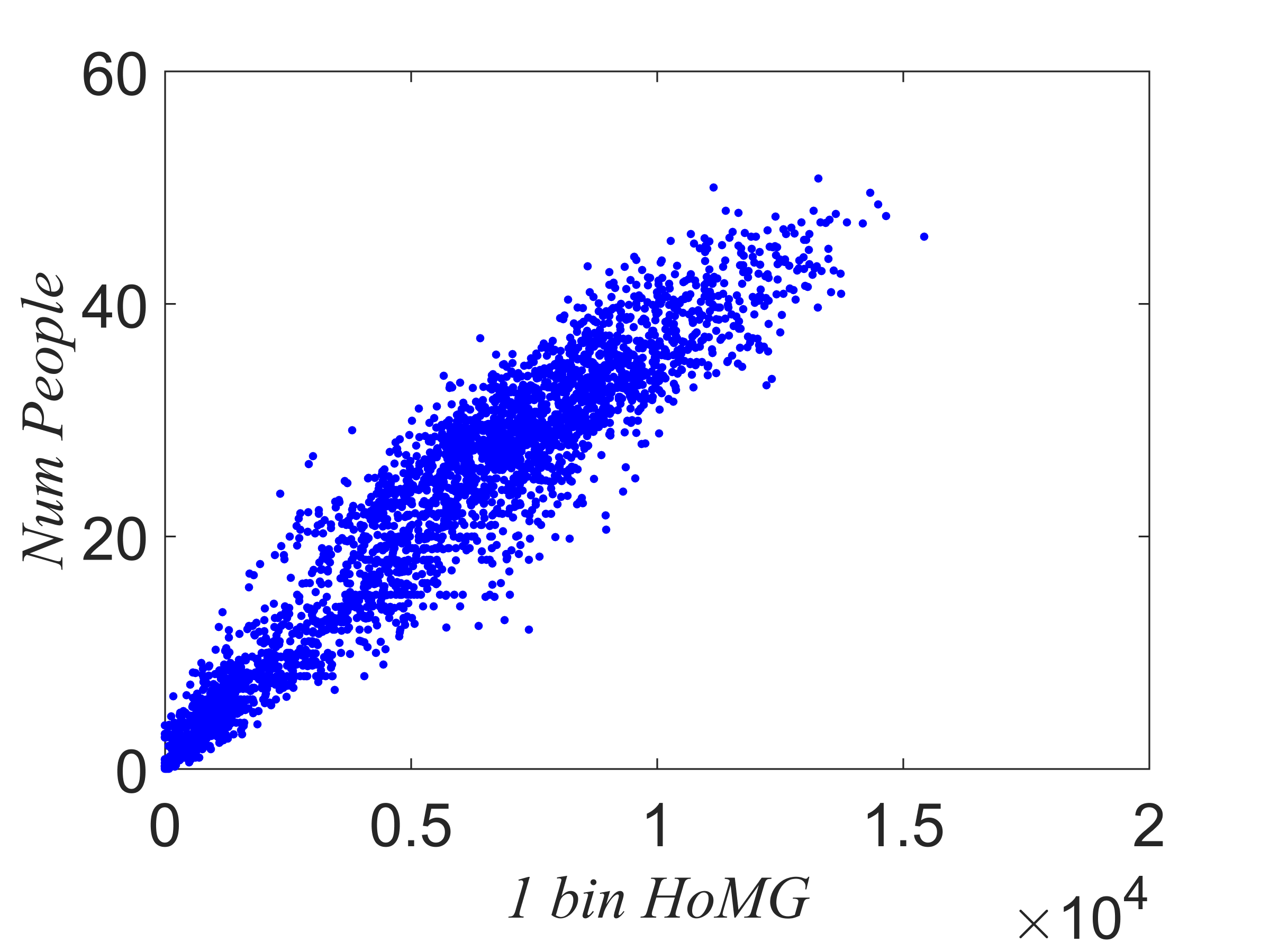}
        \caption{\small \em 13 Cameras All Data}
        \label{fig:13All}
    \end{subfigure}
    ~ 
    \begin{subfigure}[b]{0.46\linewidth}
        \includegraphics[width=\textwidth]{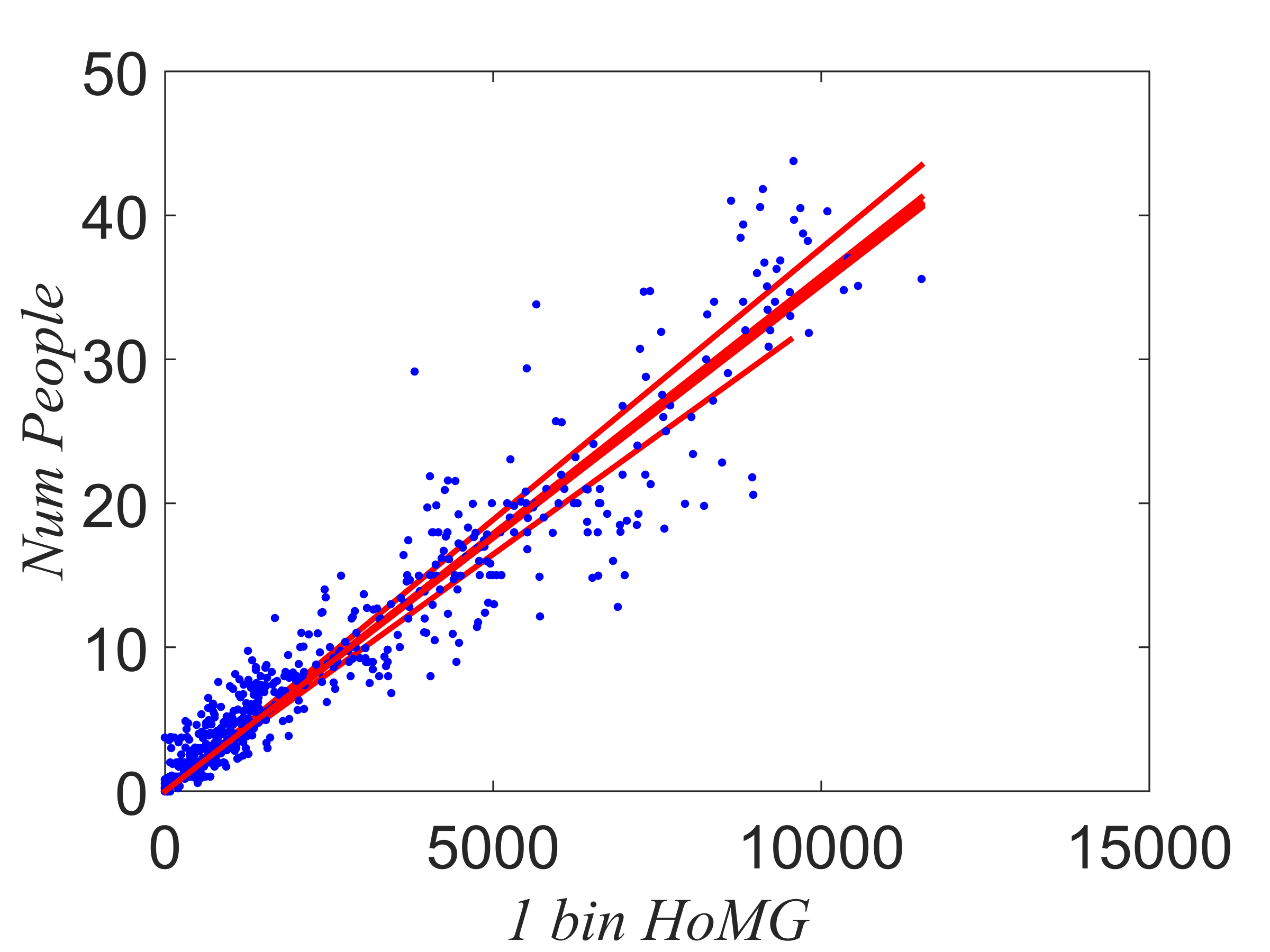}
        \caption{\small \em 13 Cameras Training Data}
        \label{fig:13fit}
    \end{subfigure}
    \end{center}
    \caption{\small \em (a) crowd-region HoMG feature vs number of people in the scene. (b) plots only the data used for training (first 50 frames) as well as all linear SVR lines obtained during leave-one-out testing on the camera views.}\label{fig:HoMGvsPeople}
\end{figure}

\begin{figure*}[t]
\begin{center}
\includegraphics[width=0.25\linewidth]{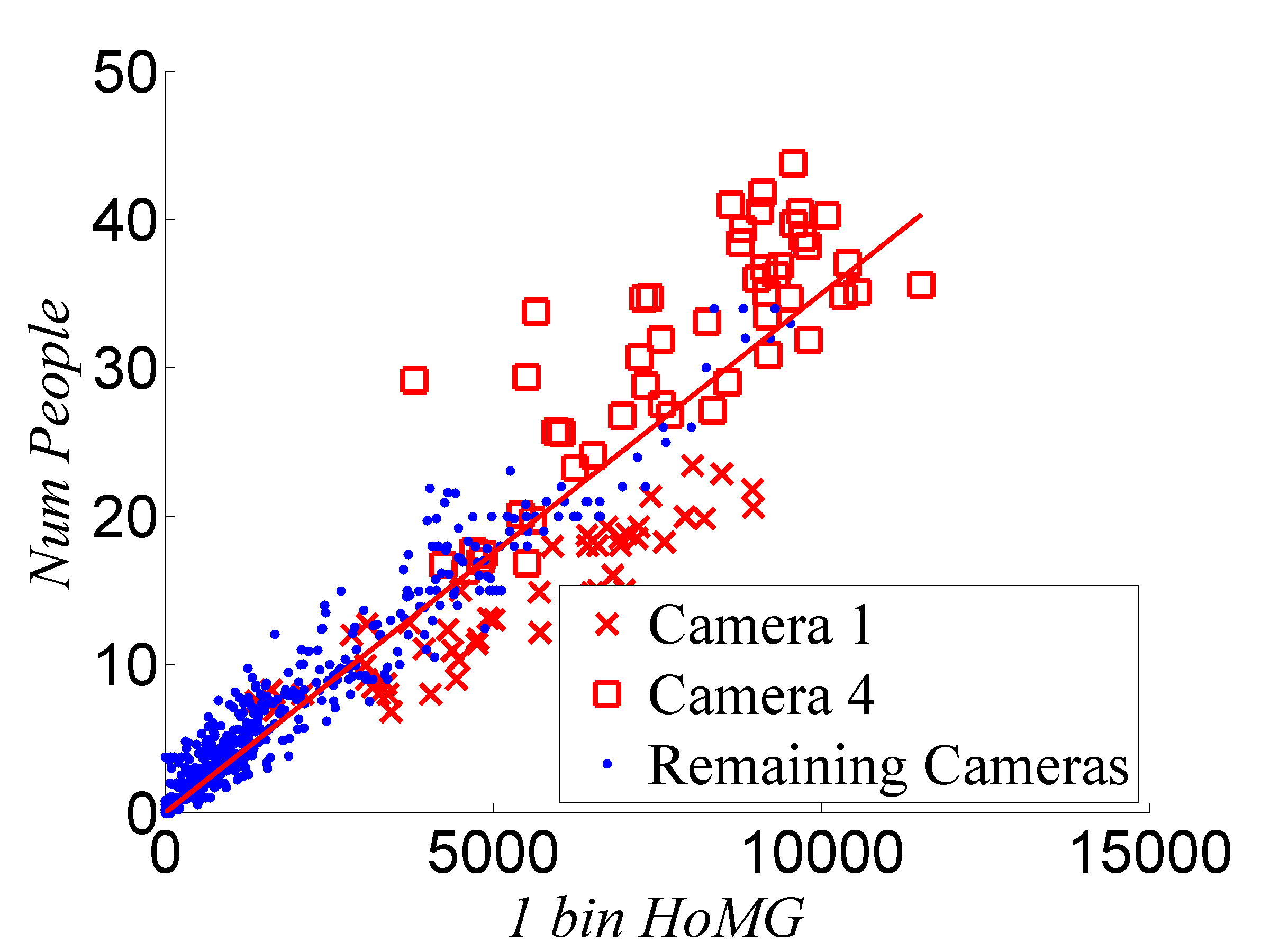}
\includegraphics[width=0.3\linewidth]{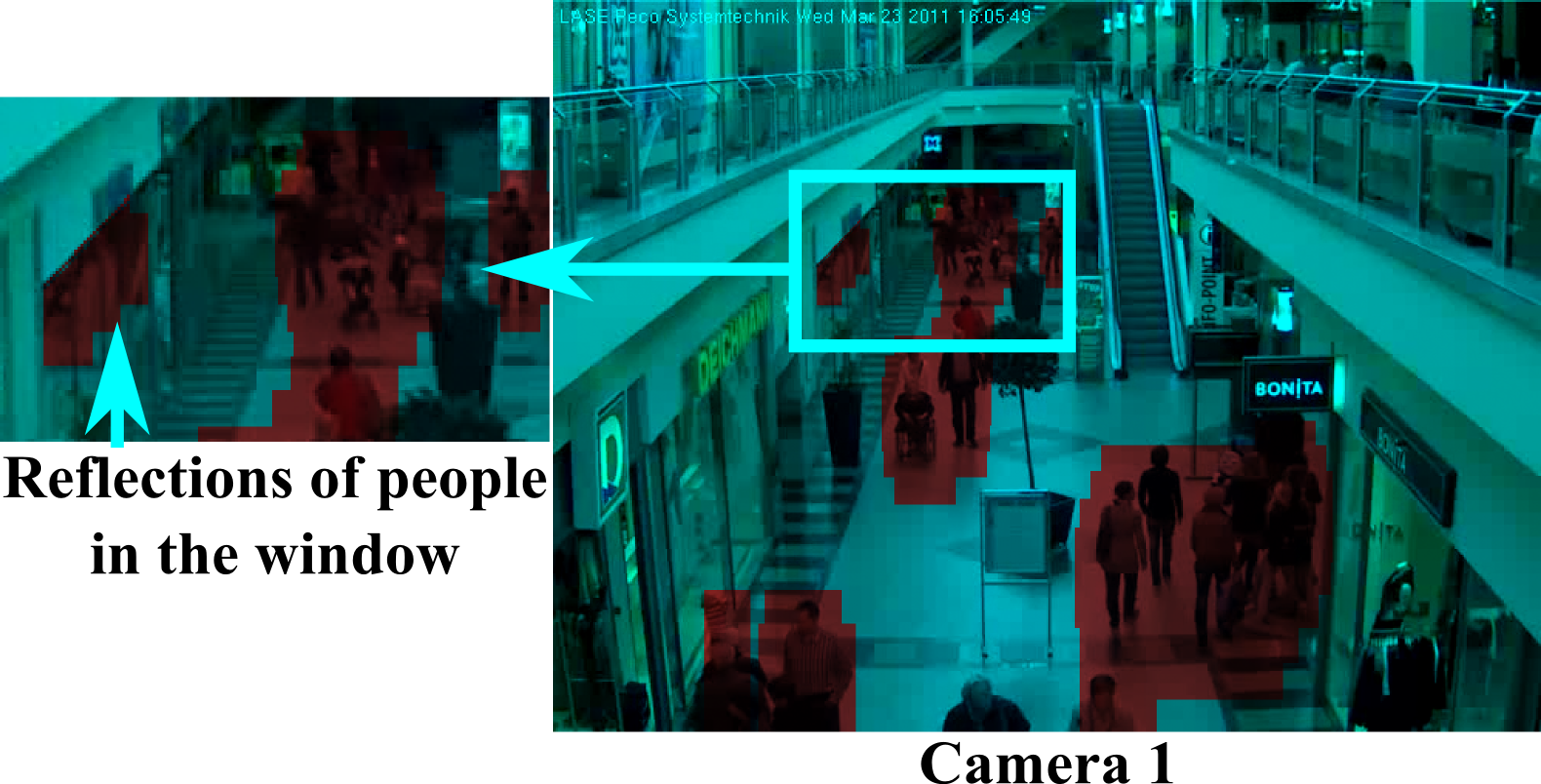}
\includegraphics[width=0.3\linewidth]{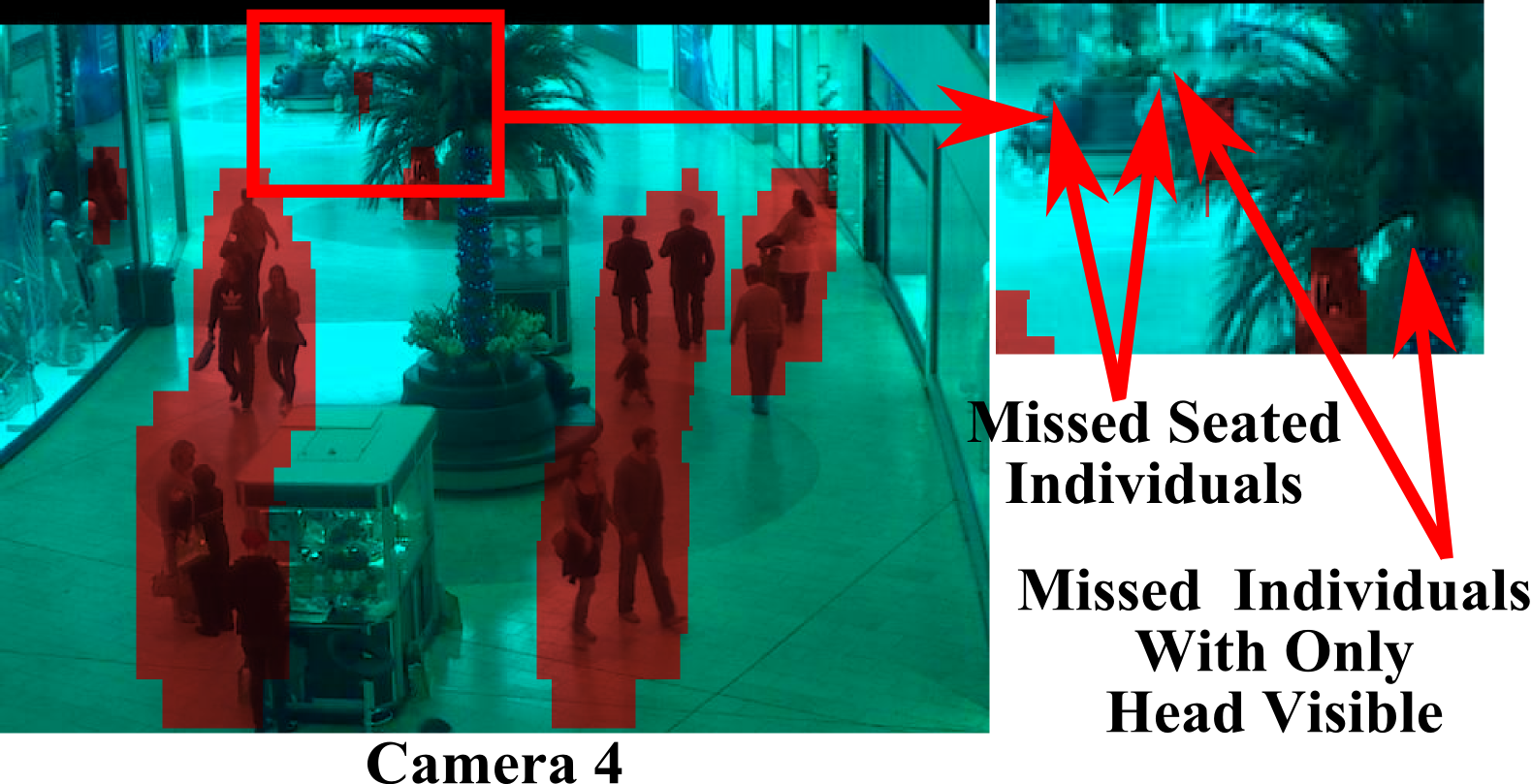}
\end{center}
   \caption{\small \em Camera 1 and 4 differs slightly from the linear SVR fit, exhibiting consistent bias below or above the linear SVR fit, respectively. This causes an over-estimation of crowd size in Camera 1 and an under-estimation in Camera 4. The over-estimation is due to edge contributions of people's reflections in the store windows and the under-estimation is due to missed detection of seated individuals.}
\label{fig:overAndUnderCounts}
\end{figure*}

To illustrate that the crowd-region HoMG feature is linearly related to the people count, we plot the crowd-region HoMG feature vs number of people in the frame in Fig.~\ref{fig:HoMGvsPeople}. The 13 regression lines (Fig.~\ref{fig:13fit}) -- obtained from leave-one-out testing -- are almost identical, indicating this is a highly-effective and consistent scene-invariant feature. The average slope of the 13 lines is $3.5\text{\sc{e}-}3$ with a standard deviation of $9.9\text{\sc{e}-}5$. However, even with this small deviation in the fitted line there are two outliers, which we take a closer look in Fig.~\ref{fig:overAndUnderCounts}. From Fig.~\ref{fig:overAndUnderCounts} we can see that for Camera 1 there is a fairly consistent over-estimation of the people count, while there is an under-estimation of the people count in Camera 4.

On closer examination of Camera 1, we find that the store front glass displays are acting as a mirror (Fig.~\ref{fig:overAndUnderCounts}), reflecting people as they walk by. The reflections are classified as crowd regions by HoMG and are thus included in our regression estimation, resulting in an over-estimation of the people count.

On closer examination of Camera 4 we find that there are missed detections due to people sitting down and only their heads being visible (Fig.~\ref{fig:overAndUnderCounts}). Seated people are very stationary for long periods of time, and as such our moving gradient based approach to crowd segmentation will tend to miss them. Our HoMG block for crowd segmentation covers a person's torso (Fig.~\ref{fig:slidingwindow}) and as such when only an individual's head is visible, HoMG will not be as effective.

\subsection{Comparison to State-Of-The-Art (7 Cameras)}
\label{subsec:datasubset}
\begin{table}
\begin{center}
\setlength{\tabcolsep}{0.5mm}
\scriptsize{
\begin{tabular}{|c|c|c|c|c|c|c|c|c||c|}
\hline
\multicolumn{2}{|c|}{\textbf{Camera}} & 7 & 8 & 9 & 10 & 11 & 12 & 13 & Avg. \\
\hline
 \rowcolor{Gray}
\multirow{3}{*}{\textbf{MAE}} & HoMG & 1.741 &  \textbf{1.534} & \textbf{ 0.303} &  \textbf{0.493} & 1.749 &  \textbf{0.849} & \textbf{ 0.330} &  \textbf{0.926} \\
&SIM3C &  \textbf{1.321} & 3.365 & 0.405 & 1.574 &  \textbf{0.886 }& 1.448 & 0.487 & 1.355 \\
 \rowcolor{Gray}
&S3C & 6.84 & 4.74 & 1.51 & 15.55 & 4.26 & 2.03 & 1.65 & 5.23 \\\hline\hline
 \rowcolor{Gray}
\multirow{3}{*}{\textbf{MSE}} & HoMG & \textbf{ 3.893} & \textbf{ 3.956} &  \textbf{0.294 }&  \textbf{0.469} & 4.031 &  \textbf{1.158} &  \textbf{0.182} &  \textbf{1.997} \\
&SIM3C & 4.250 & 17.514 & 0.495 & 3.506 &  \textbf{1.524} & 3.625 & 0.441 & 4.479 \\
 \rowcolor{Gray}
 &S3C & 76.77 & 38.154 & 5.52 & 3.91 & 50.06 & 6.61 & 3.79 & 26.40 \\\hline\hline
  \rowcolor{Gray}
 \multirow{3}{*}{\textbf{MDE}} &  HoMG & 0.123 & 0.284 & 0.363 & 0.263 & 0.186 & 0.204 &  \textbf{0.191} & 0.231 \\
 &SIM3C&  \textbf{0.103} &  \textbf{0.096} &  \textbf{0.250} & \textbf{ 0.140 }&  \textbf{0.122} &  \textbf{0.182} & 0.222 &  \textbf{0.159} \\
  \rowcolor{Gray}
 &S3C& 0.72 & 1.99 & 31.82 & 44.23 & 0.45 & 0.69 & 1.82 & 11.67\\ \hline
\end{tabular}
}
\end{center}
\caption{\small \em  Crowd counting results on 7 Camera subset used in \cite{Ryan201498}.}
\label{tab:crowdCount7Cam}
\end{table}

We compare HoMG to the state-of-the-art results reported by Ryan et al. \cite{Ryan201498} (\textbf{SIM3C}) on the same 7 Camera views used by Ryan et al. For a fair comparison, we train both the proposed crowd segmentation and regression methods only on the 7 Camera views used in \textbf{SIM3C} \cite{Ryan201498}. The results of the crowd counting are shown in Table~\ref{tab:crowdCount7Cam}.

HoMG, on average, achieved 1.5 times lower MAE and 2.25 times lower MSE than \textbf{SIM3C}. This result is achieved even though we use a single feature (HoMG) in comparison to \textbf{SIM3C} which use multiple features such as moving blob shape features, edge features, and key point descriptors. However, HoMG exhibited a 1.4 times higher MDE than \textbf{SIM3C}, due to the variation of data from the line fit seen in Fig.~\ref{fig:13All}, especially when the number of people in the frame is low.

\section{Timing Analysis}
\label{sec:TimeAnalysis}

The proposed HoMG approach was implemented using C++ and the average processing times was computed for each of the 13 Camera views using the original video resolution. Using an Intel Core i7-2720QM processor at 2.2GHz, it takes on average $18ms$ to process a frame. This is equivalent to over $50$ frames per second (FPS) in terms of processing capabilities, thus allowing for real-time crowd analysis. Further hardware specific optimization, such as SSE optimization, can yield even faster run times.

\section{Importance of Scale Normalization}

\begin{figure}[t]
\begin{center}
\includegraphics[width=0.9\linewidth]{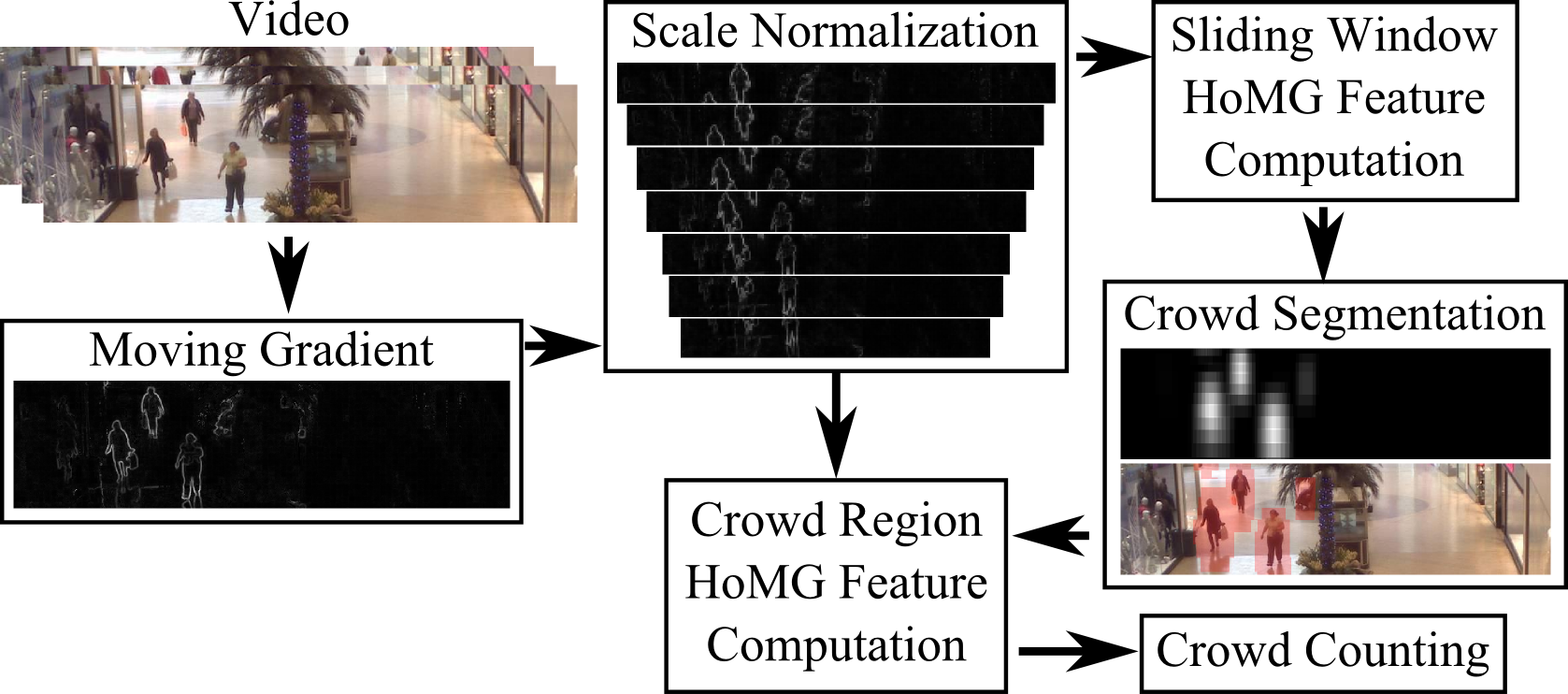}
\end{center}
   \caption{\small \em Modified version of proposed HoMG computation approach illustrated in Fig.~\ref{fig:block}, where we compute moving gradients before scale normalization.}
\label{fig:blockModified}
\end{figure}

In computing scale-normalized HoMG (Fig.~\ref{fig:block}), each section of the input frame is normalized relative to person size obtained from camera calibration prior to computing moving gradients.  This approach is similar to that of sliding-window based object detection algorithms \cite{negMine}. However, in most crowd counting literature~\cite{Loy2013,Ryan201498} the features, including edge based features, are computed on the original frame. To study the effect of computing features on original-resolution videos vs. scale- normalized representaions, we propose a slight modification to the proposed HoMG computation. As shown in Fig.~\ref{fig:blockModified}, we modify the HoMG computation algorithm by computing the moving gradients on the original-resolution frame prior to scale normalizing the moving gradients relative to person size obtained from camera calibration.

\begin{figure}[t]
\begin{center}
\includegraphics[width=0.5\linewidth]{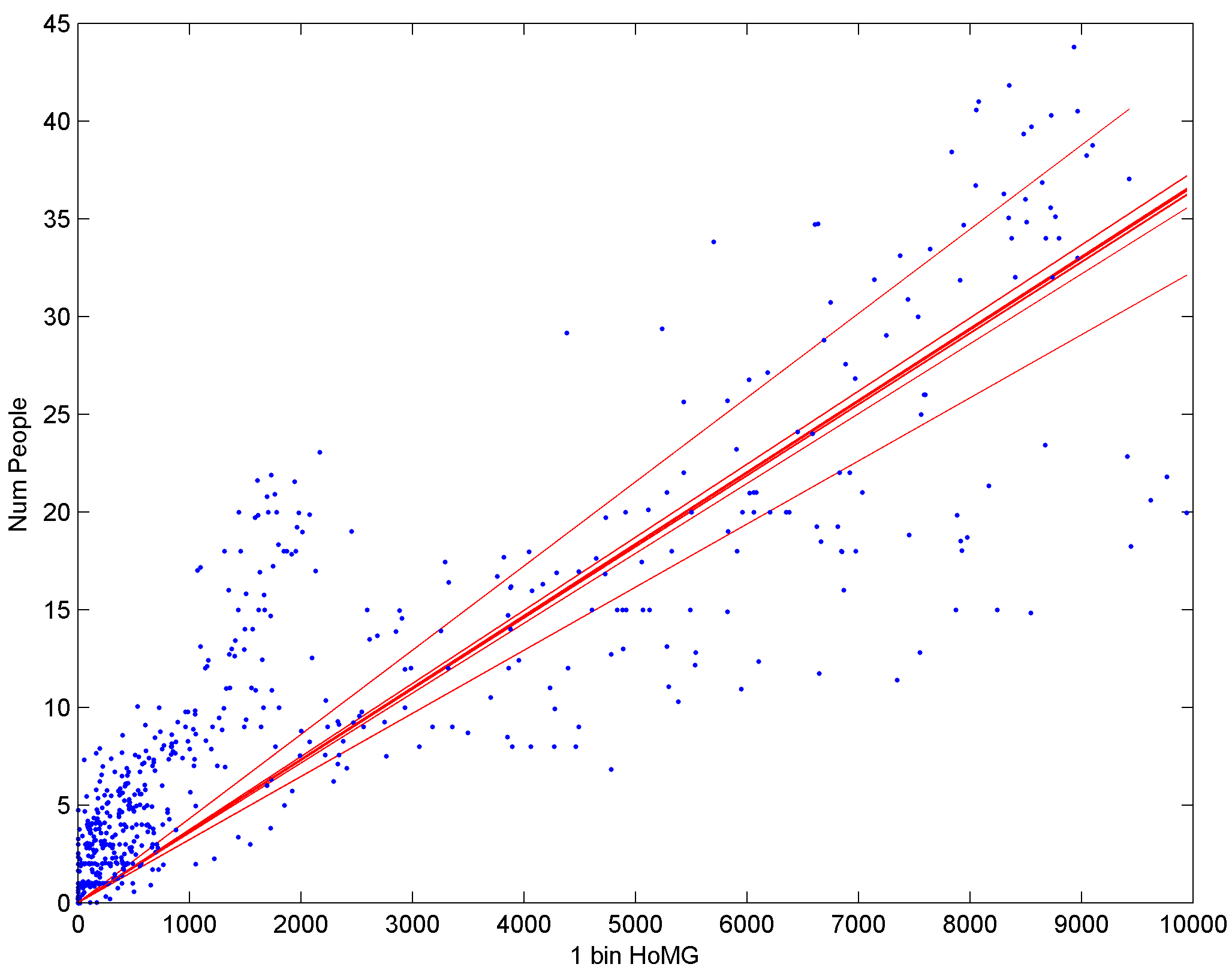}
\end{center}
    \caption{\small \em Modified crowd-region HoMG vs. people count for the training data (first 50 annotated frames) on the same axis as leave-one-out regression lines. Modified HoMG because moving gradient is computed for HoMG before scale normalization.}
\label{fig:modPlot}
\end{figure}

In our modified HoMG computation approach, moving gradients are effectively computed at different scales due to perspective effects in the image. As a result, we expect inconsistent moving gradient information during regression. In fact, this is what occurs as can be seen in Fig.~\ref{fig:modPlot} where we plot the modified crowd-region HoMG feature vs. people count for all 13 camera views. Unlike the original plot Fig.~\ref{fig:13fit}, when scale normalization isn't performed before moving gradient computation (Fig.~\ref{fig:modPlot}) we find that the modified crowd-region HoMG feature is no longer linearly related to the number of people in the scene. 

\section{Conclusions}

In this paper, a novel, low-complexity scale-normalized histogram of moving gradients (HoMG) feature is introduced for robust and fast scene-invariant crowd segmentation and counting.  Experimental results using existing multi-camera datasets demonstrate that the proposed crowd counting method using HoMG can outperform state-of-the-art approaches.  Furthermore, we also introduce an expanded dataset with 13 camera views with much greater change in camera angles to demonstrate the performance of the proposed method for both crowd segmentation and counting. Based on the existing and expanded datasets, we show that the proposed method using HoMG facilitates for robust, real-time crowd analysis, which is important for widespread industrial adoption. 

{\small
\bibliographystyle{ieee}
\bibliography{crowdPapers}
}

\end{document}